\documentclass[conference]{IEEEtran}
\IEEEoverridecommandlockouts
\usepackage{subfigure} 
\usepackage{cite}
\usepackage{amsmath,amssymb,amsfonts}
\usepackage{algorithmic}
\usepackage{graphicx}
\usepackage{textcomp}
\usepackage{xcolor}
\usepackage{mathrsfs}
\usepackage{marvosym}
\IEEEoverridecommandlockouts
\begin{document}

\title{Multiple Lane Detection Algorithm Based on Optimised Dense Disparity Map Estimation}

\author{Han Ma, Yixin Ma, Jianhao Jiao, M Usman Maqbool Bhutta, \\Mohammud Junaid Bocus, Lujia Wang, Ming Liu, Rui Fan$^*$

\thanks{

	H. Ma is with the Department of Precision Instrument, School of Mechanical Engineering, Tsinghua University, Beijing, China. (e-mail: mh15@mails.tsinghua.edu.cn)
	
	Y. Ma is with the School of Electronic Information and Electrical Engineering, Shanghai Jiao Tong University, Shanghai, China. (e-mail: y.ma@sjtu.edu.cn)
	
	J. Jiao is with the Robotics and Multi-Perception Laborotary, Robotics Institute, The Hong Kong University of Science and Technology, Hong Kong SAR, China. (e-mail: jjiao@connect.ust.hk)
	
	M. U. M. Bhutta is with the Robotics and Multi-Perception Laborotary, Robotics Institute, The Hong Kong University of Science and Technology, Hong Kong SAR, China. (e-mail: mumbhutta@connect.ust.hk)
	
	M. J. Bocus is with the Visual Information Group, the University of Bristol, BS8 1UB, UK. (e-mail: junaid.bocus@bristol.ac.uk)
	
	L. Wang is with the Shenzhen Institutes of Advanced Technology, Chinese Academy of Sciences, Shenzhen, China. (e-mail: lj.wang1@siat.ac.cn)
	
	M. Liu is with the Robotics and Multi-Perception Laborotary, Robotics Institute, The Hong Kong University of Science and Technology, Hong Kong SAR, China. (e-mail: eelium@ust.hk)
	
	R. Fan is with the Robotics and Multi-Perception Laborotary, Robotics Institute, The Hong Kong University of Science and Technology, Hong Kong SAR, China. (e-mail: rui.fan@ieee.org)
	
{*Corresponding author: Rui Fan.}
	
}}

\maketitle

\begin{abstract}
Lane detection is very important for self-driving vehicles. In recent years, computer stereo vision has been prevalently used to enhance the accuracy of the lane detection systems. This paper mainly presents a multiple lane detection algorithm developed based on optimised dense disparity map estimation, where the disparity information obtained at time $\boldsymbol{t_n}$ is utilised to optimise the process of disparity estimation at time $\boldsymbol{t_{n+1}}$ ($\boldsymbol{n\geq0}$). This is achieved by 
estimating the road model at time $\boldsymbol{t_{n}}$ and then controlling the search range for the disparity estimation at time $\boldsymbol{t_{n+1}}$. The lanes are then detected using our previously published algorithm, where the vanishing point information is used to model the lanes. The experimental results illustrate that the runtime of the disparity estimation is reduced by around $\boldsymbol{37\%}$ and the accuracy of the lane detection is about $\boldsymbol{99\%}$. 
 
\end{abstract}

\begin{IEEEkeywords}
lane detection, self-driving vehicles, stereo vision, disparity estimation, vanishing point.
\end{IEEEkeywords}

\section{Introduction}
\label{sec.introduction}
\IEEEPARstart{T}HE deployment of autonomous vehicles has been increasing rapidly since Google first launched their self-driving car project in 2009 \cite{Brink2017}. In recent years, with a number of technology breakthroughs being witnessed in the world where science fiction inventions are now becoming a reality, the competition to commercialise driver-less vehicles by companies like GM, Waymo and Daimler-Bosch is fiercer than ever \cite{Fan2017}. For instance, Volvo conducted a series of self-driving experiments involving around 100 cars in China \cite{Fan2016}. The 5G network is also utilised in self-driving vehicles to help them communicate with each other \cite{Fan2017}. Furthermore, the computer stereo vision technique has also been prevalently used in prototype vehicle road tests to provide the depth information for various automotive applications, e.g., road condition assessment \cite{Fan2018, Fan2018c}, lane and obstacle detection \cite{Fan2018a, Fan2016}, and vehicle state estimation \cite{Evans2018a,Fan2018d}.

The state-of-the-art lane detection algorithms can mainly be classified as feature-based and model-based \cite{Fan2018a}. The former extracts local and useful information of an image, such as edges, texture and colour, for the segmentation of  lanes and road boundaries \cite{Bertozzi1998}. On the other hand, the model-based algorithms aim to represent the lanes with mathematical equations, e.g., linear, parabolic, linear-parabolic and spline, based upon some common road geometry assumptions \cite{Wang2004}. The linear model works well for lanes with a low curvature \cite{Fan2016}, but a more flexible road model is inevitable when lanes with a higher curvature are considered. Therefore, some algorithms \cite{Kluge1995, Wang2008, Zhou2006, Kreucher1999} use a parabolic model to represent the lanes with a constant curvature. To address some more complex cases, Jung et al. proposed a combination of linear and parabolic lane models, where the far lanes are modelled as parabolas whereas the nearby lanes are represented as linear models \cite{Jung2005}. In addition to the models mentioned above, the spline model is an alternative method whereby the lane pixels are interpolated into an arbitrary shape \cite{Wang2004, Wang2000}. However, the more parameters introduced into a flexible model, the higher will be the  computational complexity of the algorithm. Therefore, some authors resort to additional important properties of 3-D imaging techniques, such as disparity estimation \cite{Fan2018a}, instead of being limited to only 2-D information.

\begin{figure}[!t]
	\begin{center}
		\centering
		\includegraphics[width=0.48\textwidth]{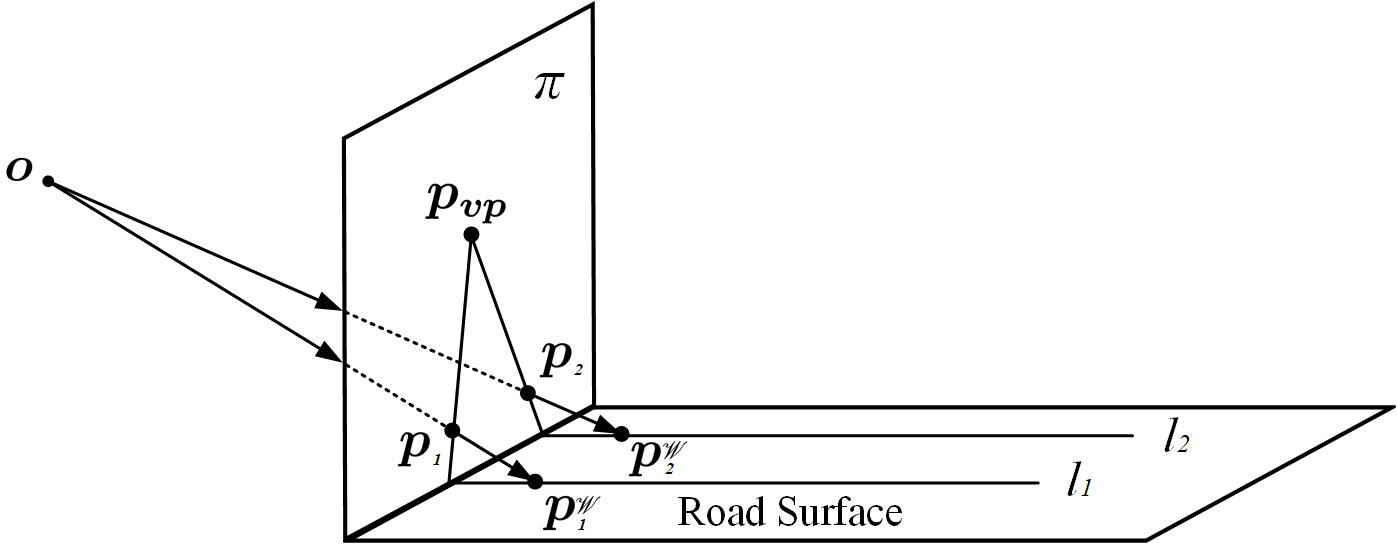}
		\caption{Vanishing point.}
		\label{fig.vanishing_point}
	\end{center}
	\vspace{-1em}
\end{figure}

\begin{figure}[!t]
	\begin{center}
		\centering
		\includegraphics[width=0.40\textwidth]{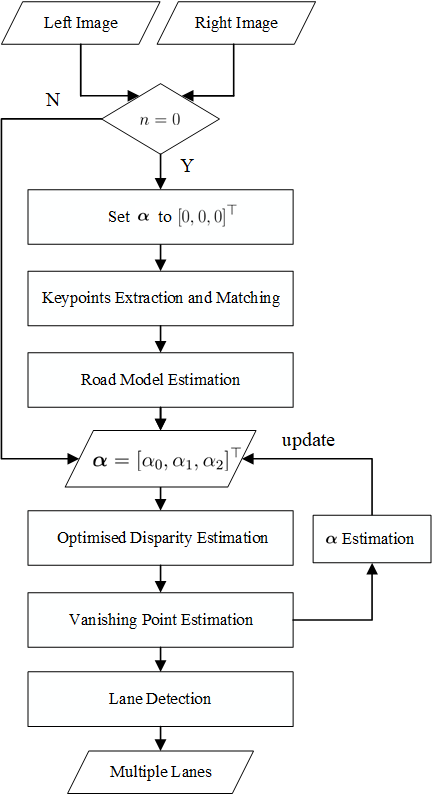}
		\caption{Flow chart of the proposed algorithm.}
		\label{fig.flow_chart}
	\end{center}
	\vspace{-1em}
\end{figure}

One of the most prevalently used 3-D based methods is Inverse Perspective Mapping (IPM). With the assumption that two lanes are parallel to each other in the World Coordinate System (WCS), IPM can map a 3-D scenery into a 2-D bird's eye view \cite{Nieto2007}. In addition, some authors \cite{Schreiber2005, Hanwell2012, Fardi2004, Ozgunalp2017} proposed to use the vanishing point $\boldsymbol{{p}_{vp}}=[u_{vp}, v_{vp}]^\top$ to model lane markings and road boundaries, where $u_{vp}$ and $v_{vp}$ denote the vertical and horizontal coordinates of the vanishing point, respectively. An example of vanishing point is illustrated in Fig. \ref{fig.vanishing_point}, where $l_1$ and $l_2$ are two parallel straight lines. Two 3-D points $\boldsymbol{p_1}^\mathscr{W}$ and $\boldsymbol{p_2}^\mathscr{W}$ in the WCS are projected on the image plane $\pi$ and their corresponding points in the Image Coordinate System (ICS) are $\boldsymbol{p_1}$ and $\boldsymbol{p_2}$, respectively. Therefore, the projections of $l_1$ and $l_2$ in the image are two straight lines  and they intersect at the vanishing point $\boldsymbol{{p}_{vp}}$. However, their algorithms work well only if the road surface is assumed to be flat or the camera parameters are known. Hence, some researchers pay closer attention to the disparity information which can be obtained using either active sensors, e.g., radar and laser, or passive sensors, e.g., stereo cameras \cite{Ozgunalp2017}. Since Labayrade et al. proposed the concept of \textquotedblleft v-disparity" in 2002 \cite{Labayrade2002}, disparity information has been widely used to enhance the robustness of the lane detection systems. The  work presented in \cite{Ozgunalp2017} shows a particular instance where the disparity information is successfully combined with a lane detection algorithm to estimate $\boldsymbol{{p}_{vp}}$ for both flat and non-flat road surfaces. At the same time, the obstacles containing a lot of redundant information can be eliminated by comparing the actual and fitted disparity values. However, the estimation of $\boldsymbol{{p}_{vp}}$ is greatly affected by the outliers when performing the Least Squares Fitting (LSF), and the lanes are sometimes wrongly detected because the selection of plus-minus peaks is not always effective. Moreover, achieving real-time performance is still challenging in \cite{Ozgunalp2017} because of the intensive computational complexity of the algorithm. Hence in this work, we propose a more efficient lane detection algorithm where the disparity information 
acquired at time $t_{n-1}$ is utilised to improve the process of the disparity estimation at time $t_{n}$. The flow chart of the proposed lane detection algorithm is depicted in Fig. \ref{fig.flow_chart}. 

Firstly, we create a vector $\boldsymbol{\alpha}=[\alpha_0,\alpha_1,\alpha_2]^\top$ to store the parameters of the road model. The keypoints $\boldsymbol{P}=[\boldsymbol{p_0}, \boldsymbol{p_1},\dots, \boldsymbol{p_n}]^\top$  and $\boldsymbol{Q}=[\boldsymbol{q_0}, \boldsymbol{q_1},\dots, \boldsymbol{q_n}]^\top$ are then extracted and matched using Binary Robust Invariant Scalable Keypoints (BRISK), where $\boldsymbol{p_n}=[u_{pn},v_{pn}]^\top$ and  $\boldsymbol{q_n}=[u_{qn},v_{qn}]^\top$ denote the matched keypoints in the left and right images, respectively. $\boldsymbol{P}$ and $\boldsymbol{Q}$ are then utilised to create a sparse v-disparity histogram from which the road model $f(v)=\alpha_0+\alpha_1v+\alpha_2v^2$ can be estimated. The search range for the disparity estimation at time $t_0$ is subsequently controlled according to $f(v)$, where the search range at row $v$ is limited to $[f(v)-\tau,f(v)+\tau]$. The lanes are then detected using our previously published method in \cite{Ozgunalp2017}. As for the disparity estimation at time $t_{n+1}$ ($n\geq0$), $\boldsymbol{\alpha}$ is estimated by creating a dense v-disparity map using the disparity information acquired at time $t_{n}$ ($n\geq0$) and then fitting a parabola to the best path extracted from the dense v-disparity map. This not only improves the accuracy of the disparity estimation at time $t_{n+1}$ but also boosts its processing speed. 

The remainder of the paper is structured as follows: Section \ref{sec.algorithm_description} presents the proposed lane detection algorithm. In Section \ref{sec.experimental_results}, the experimental results are illustrated and the evaluation of the proposed system is carried out. Finally, Section \ref{sec.conclusion_future_work} summaries the paper and provides some recommendations for future work.

\section{Algorithm Description}
\label{sec.algorithm_description}

\subsection{Disparity Estimation}
\label{sec.disparity_estimation}

The proposed disparity estimation approach in this paper is developed based on our previously published algorithm \cite{Fan2017}, where the search range at the position of $(u,v)$ is propagated from three estimated disparities at the positions of $(u-1,v)$, $(u,v)$ and $(u+1,v)$. In this subsection, we optimise the process of the disparity estimation by utilising the disparity information acquired at time $t_{n-1}$ to control the search range at time $t_{n}$, where $n$ is a positive integer.
 
\begin{figure*}[!t]
	\begin{center}
		\centering
		\subfigure[]
	{\includegraphics[width=0.74\textwidth]{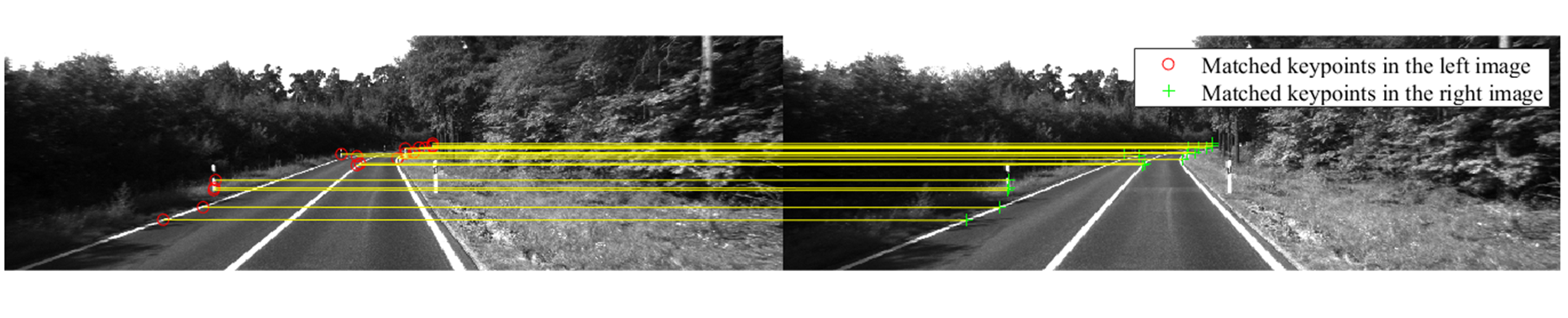}
	\label{fig.feature_matching1}
}
\subfigure[]
{		\includegraphics[width=0.22\textwidth]{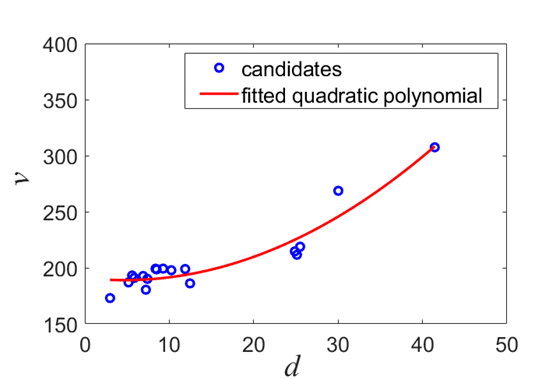}
	\label{fig.feature_matching2}
}
		\caption{Road model estimation for the initial stereo image pair. (a) BRISK-based on-road keypoints extraction and matching between the left and right images. (b) the corresponding sparse v-disparity histogram of Fig. \ref{fig.feature_matching1}. }
		\label{fig.feature_matching}
			\vspace{-1em}
	\end{center}
\end{figure*}

\begin{figure}[!t]
	\centering
	\subfigure[]
	{
		\includegraphics[width=0.38\textwidth]{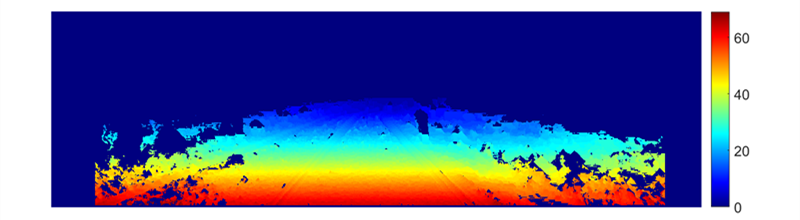}
		\label{fig.disp0}
	}
	\subfigure[]
	{
		\includegraphics[width=0.38\textwidth]{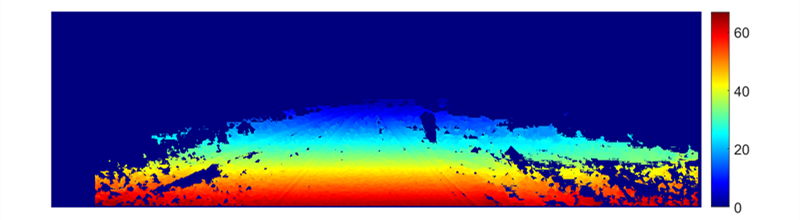}
		\label{fig.disp1}
	}
	\subfigure[]
	{
		\includegraphics[width=0.38\textwidth]{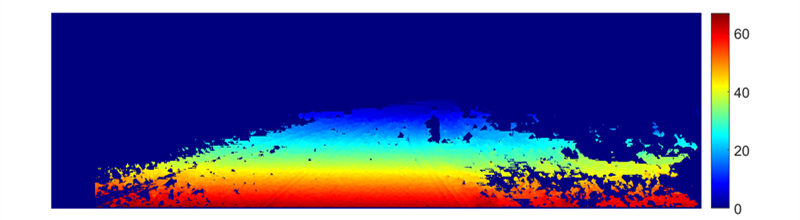}
		\label{fig.disp38}
	}
	\caption{Disparity map estimation at time $t_{n}$. (a) $n=0$. (b) $n=1$. (c) $n=38$.}
	\label{fig.disp_estimation}
		\vspace{-1em}
\end{figure}

As for the stereo matching at time $t_{0}$, the parameter vector $\boldsymbol{\alpha}=[\alpha_0,\alpha_1,\alpha_2]^\top$ is first initialised as $[0,0,0]^\top$. Then, $\boldsymbol{\alpha}$ is estimated by solving a least squares problem with a set of reliable correspondences $\boldsymbol{P}=[\boldsymbol{p_0}, \boldsymbol{p_1},\dots, \boldsymbol{p_n}]^\top$  and $\boldsymbol{Q}=[\boldsymbol{q_0}, \boldsymbol{q_1},\dots, \boldsymbol{q_n}]^\top$. In this paper, we use BRISK to detect and match $\boldsymbol{P}$ and $\boldsymbol{Q}$. It allows a faster execution to achieve approximately the same number of correspondences as SURF (Speeded-Up Robust Features) and SIFT (Scale-Invariant Feature Transform) \cite{Leu2011}. An example of keypoints detection and matching is shown in Fig. \ref{fig.feature_matching1}. The correspondences are then utilised to create a sparse v-disparity histogram as shown in Fig. \ref{fig.feature_matching2}. The values of $\alpha_0$, $\alpha_1$ and $\alpha_2$ can then be obtained by fitting a parabola to the non-zero candidates in the sparse v-disparity map. Since the outliers can severely affect the accuracy of the LSF, Random Sample Consensus (RANSAC) is utilised to 
iteratively update $\boldsymbol{\alpha}$ until the ratio $r=n_{\mathcal{I}}/(n_{\mathcal{O}}+n_{\mathcal{I}})$ reaches a maximum value, where $n_{\mathcal{I}}$ and $n_{\mathcal{O}}$ represent the number of inliers and outliers, respectively. Then, the search range at row $v$ is limited to $[f(v)-\tau,f(v)+\tau]$, where $\tau$ is a threshold set to $3$. The estimated disparity map at time $t_0$ is shown in Fig. \ref{fig.disp0}. 

Afterwards, the disparity information acquired at time $t_{n-1}$ $(n>0)$ is utilised to create a dense v-disparity map. The latter is then optimised using Dynamic Programming (DP) and the optimum solution corresponds to the vertical road profile. More details are provided in subsection \ref{sec.lane_detection}.  The optimum solution is then interpolated into a quadratic polynomial and the values of $\alpha_0$, $\alpha_1$ and $\alpha_2$ at time $t_{n-1}$ can be obtained. The parameter vector $\boldsymbol{\alpha}$ at time $t_{n-1}$ is then used to control the search range at time $t_{n}$ according to the strategy mentioned above. The estimated disparity maps at time $t_1$ and $t_{38}$ are illustrated in Fig. \ref{fig.disp1} and Fig. \ref{fig.disp38}, respectively.

\subsection{Image Segmentation and Edge Detection}
\label{sec.image_segmentation}

\subsubsection{Bilateral Filtering}
\begin{figure}[!t]
	\centering
	\subfigure[]
	{
		\includegraphics[width=0.30\textwidth]{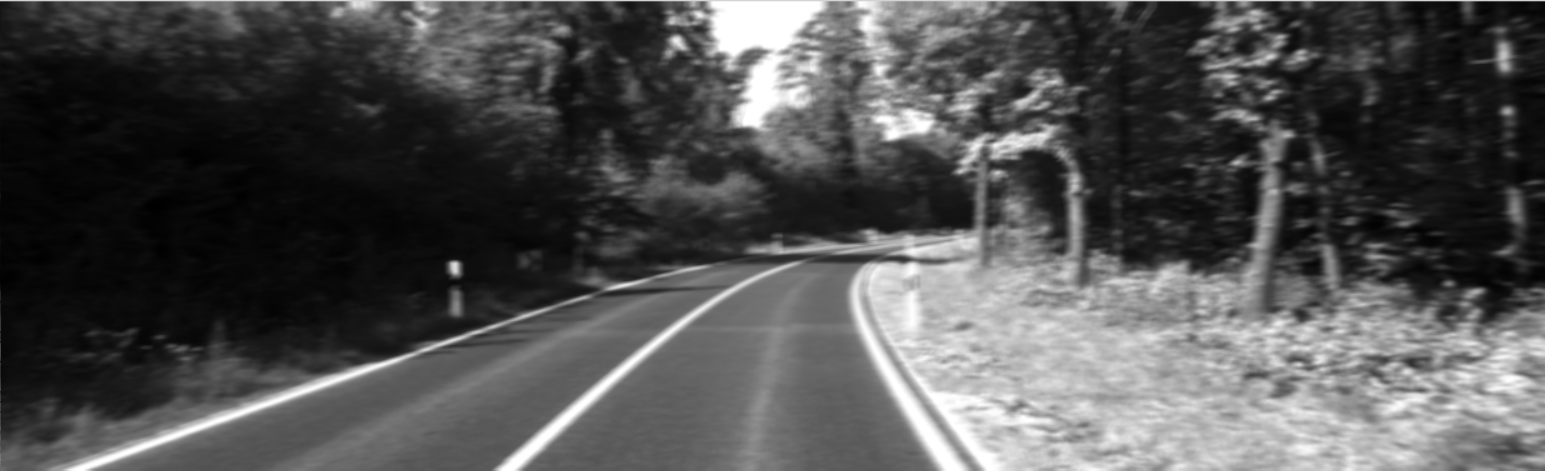}
		\label{fig.bilateral_filtering}
	}
	\subfigure[]
	{
		\includegraphics[width=0.30\textwidth]{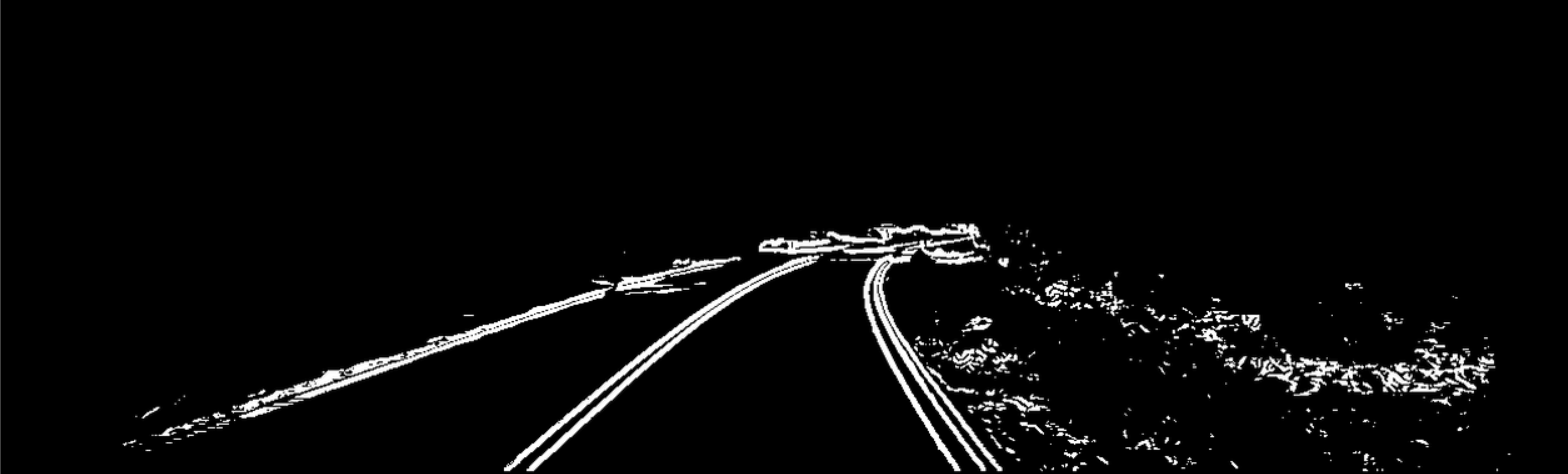}
		\label{fig.edge_detection}
	}
	\caption{Bilateral filtering and edge detection. (a) the result of bilateral filtering. (b) the edge detection result of Fig. \ref{fig.bilateral_filtering}. }
	\label{fig.bilateral_filtering_edge_detection}
	\vspace{-1em}
\end{figure}

The input image is always noisy, making various edge detectors such as Sobel and Canny very sensitive to the blobs \cite{Fan2018a}. Therefore, we first perform bilateral filtering on the input image to reduce redundant information while still preserving the edges. The filtered image is shown in Fig. \ref{fig.bilateral_filtering}.

\subsubsection{Road Model Estimation}
\label{sec.road_model_estimation}

As discussed in subsection \ref{sec.lane_detection}, the boundary between the road and sky $v=v_{vp}|_{min}$ can be obtained by solving a quadratic function $f(v)=0$. Furthermore, the road area can be extracted by computing the difference between the actual and fitted disparity values, i.e., $\tilde{d}$ and $f(v)$, and then finding the pixels that meet the requirement of $\tilde{d}(u,v)\in[\max\{0, f(v)-\mu\},f(v)+\mu]$, where $\mu$ is a threshold set to remove the pixels on the obstacles \cite{Fan2018a}. 

\subsubsection{Edge Detection}
\label{sec.edge_detection}
In this subsection, we use Sobel operator to detect the edge information by convolving the horizontal and vertical Sobel operators with the original image \cite{Fan2016}. The corresponding edge detection result is shown in Fig. \ref{fig.edge_detection}.

\subsection{Lane Detection}
\label{sec.lane_detection}

DP is an optimisation method which aims at solving a complicated problem by breaking it down into a series of simpler sub-problems \cite{Fan2018a}. 
According to our previously published algorithm in \cite{Ozgunalp2017}, DP is performed to minimise the energy function in Eq. \ref{eq.energy_model}, where $E$ represents the total energy and the path with the minimum $E$ is selected as the optimum solution. $E_{data}$ depends on the total votes each accumulator gets and $E_{smooth}$ with a smoothness constant $\lambda$, penalises the changes in $\boldsymbol{{p}_{vp}}$.
\begin{equation}
E=E_{data}+\lambda E_{smooth}
\label{eq.energy_model}
\end{equation}

\begin{figure}[!t]
	\centering
	\subfigure[]
	{\includegraphics[width=0.46\textwidth]{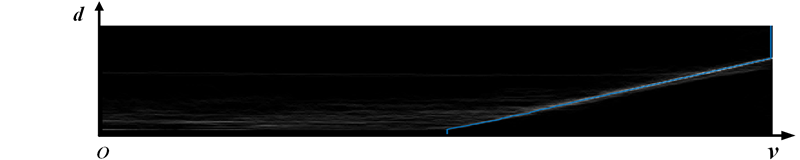}
		\label{fig.dense_v_disp_map}
	}
	\subfigure[]
	{		\includegraphics[width=0.46\textwidth]{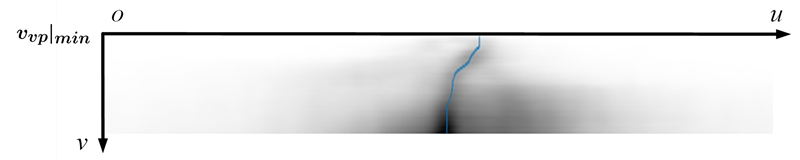}
		\label{fig.dense_u_vp_acu}
	}
	\caption{Dense $\boldsymbol{{p}_{vp}}$ estimation. (a) dense v-disparity map. The path in blue represents $f(v)$. (b) dense $u_{vp}$ map. The path in blue represents $g(v)$. }
	\label{fig.p_vp_estimation}
	\vspace{-1em}
\end{figure}

In this section, DP is first utilised to optimise the v-disparity map by minimising the energy function in Eq. \ref{eq.vpy_estimation}. The blue path in the v-disparity map is the optimum solution, as shown in Fig. \ref{fig.dense_v_disp_map}. The optimum solution is then interpolated into a quadratic polynomial: $f(v)=\alpha_0+\alpha_1v+\alpha_2v^2$, and $v_{vp}$ with respect to each row can thus be computed.

\begin{equation}
\begin{split}
&E(v)_d=-m_{v}(d,v)
\\& +\min_{\tau_v}[E(v-\tau_{v})_{d+1}-\lambda_{v}\tau_{v}],\ \text{s.t.} \  \tau_v\in[0,6]
\end{split}
\label{eq.vpy_estimation}
\end{equation}

Then, DP is used to optimise the dense $u_{vp}$ accumulator by minimising the energy function in Eq. \ref{eq.dense_vpx_estimation}.  The blue path in the dense $u_{vp}$ accumulator map is the optimum solution, as shown in Fig \ref{fig.dense_u_vp_acu}. $u_{vp}$ with respect to each row can then be estimated by interpolating the optimum solution into a quadruplicate polynomial: $g(v)=\beta_0+\beta_1v+\beta_2v^2+\beta_3v^3+\beta_4v^4$.

\begin{equation}
\begin{split}
&E(u)_v=
m_{u}(u,v) \\&+\min_{\tau_u}[E(u_{vp}+\tau_u)_{v+1}+\lambda_{u}\tau_u], \ \text{s.t.} \  \tau_u\in[-5,5]
\end{split}
\label{eq.dense_vpx_estimation}
\end{equation} 

$\boldsymbol{{p}_{vp}}$ provides the tangential direction and the curvature information of lanes, which can help to validate the lane positions. In our previous paper \cite{Ozgunalp2017}, a likelihood function $V(\boldsymbol{p_e})=\nabla(\boldsymbol{p_e})\cdot \cos(\theta_{\boldsymbol{p_e}}-\theta_{\boldsymbol{p_{vp}}})$ is formed for each edge point $\boldsymbol{{p}_{e}}$ and  the plus-minus peak pairs are selected for lane visualisation, where $\theta_{\boldsymbol{p_e}}$ is the angle between the $u$-axis and the orientation of the edge point $\boldsymbol{p_e}$, and $\theta_{\boldsymbol{p_{vp}}}$ is the angle between the $u$-axis and the radial from an edge pixel $\boldsymbol{p_e}$ to $\boldsymbol{p_{vp}}(v_e)$. More details are provided in \cite{Fan2018b, Ozgunalp2017}. The lanes can thus be visualised using $f(v)$, $g(v)$ and $V(\boldsymbol{p_e})$. Some examples of lane detection results are shown in Fig. \ref{fig.exp_results}. 

\section{Experimental Results}
\label{sec.experimental_results}

\begin{figure*}[!t]
	\begin{center}
		\centering
		\includegraphics[width=0.90\textwidth]{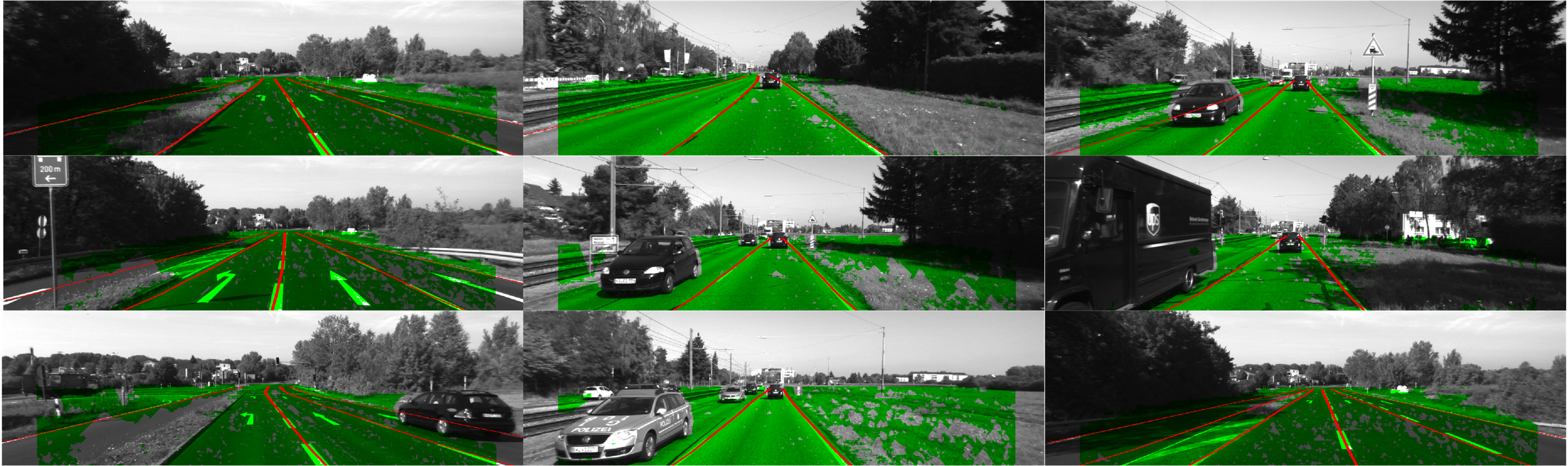}
		\caption{Experimental results. The regions in green are  estimated road surface areas. The lines in red are  detected lanes. }
		\label{fig.exp_results}
	\end{center}
	\vspace{-2em}
\end{figure*}

In this section, we use the KITTI dataset to quantify the robustness of the proposed algorithm.
Some experimental results are shown in Fig. \ref{fig.exp_results}, where the green regions are the estimated road areas and the lines in red are the detected lanes.

Currently, it is impossible to access a satisfying ground truth dataset for the evaluation of lane detection algorithms because accepted test protocols do not usually exist \cite{Hillel2014}. Therefore, many publications related to lane detection only compare the quality of their experimental results with the ones obtained using some other published algorithms \cite{Ozgunalp2017}. For this reason, we compare the performance of the proposed system with \cite{Ozgunalp2017}. 
Table \ref{table.proposed_results} and Table \ref{table.umar_detection_results} details the successful detection rates of the proposed algorithm and the algorithm presented in \cite{Ozgunalp2017}, respectively. The overall successful detection rate is around $99\%$ and it is increased by around $0.6\%$.

In addition, the proposed algorithm is implemented in C programming language. The runtime of the algorithm for a single frame with a resolution of $1242\times375$ is 0.21 seconds (using a single thread of Intel Core I7-8700K CPU). By controlling the search range at time $t_{n}$ using the disparity information acquired at time $t_{n-1}$, the runtime of the disparity estimation is reduced by around $37\%$. Although the algorithm does not perform in real time, we believe that the execution speed of the implementation can be further accelerated by highly exploiting the parallel computing architecture.

\begin{table}[!t]
	\begin{center}
		\caption{Detection results of the proposed algorithm.}
		\label{table.proposed_results}
		\vspace{1em}
		\footnotesize
		\begin{tabular}{|c|c|c|c|c|}
			\hline
			Sequence & Lanes  & Incorrect detection & Misdetection\\
			\hline
			1 & 860& 0 &  0 \\
			\hline
			2  & 594& 0 & 0 \\
			\hline
			3 & 376 &  0 & 0\\
			\hline
			4  & 156 &  0 & 3\\
			\hline
			5 & 678 &  0 & 9\\
			\hline
			6 & 1060 & 10 & 2\\
			\hline
			Total & 3724&  10 & 14 \\			
			\hline
		\end{tabular}
	\end{center}
\end{table}

\begin{table}[!t]
	\begin{center}
		\caption{Detection results of \cite{Ozgunalp2017}.}
		\label{table.umar_detection_results}
		\vspace{1em}
		\footnotesize
		\begin{tabular}{|c|c|c|c|c|}
			\hline
			Sequence & Lanes  & Incorrect detection & Misdetection\\
			\hline
			1 & 860& 0 &  0 \\
			\hline
			2  & 594& 0 & 0 \\
			\hline
			3 & 376 &  0 & 0\\
			\hline
			4  & 156 &  0 & 9\\
			\hline
			5 & 678 &  0 & 17\\
			\hline
			6 & 1060 & 14 & 7\\
			\hline
			Total & 3724&  14 & 33 \\			
			\hline
		\end{tabular}
	\end{center}
\end{table}

\section{Conclusion and Future Work}
\label{sec.conclusion_future_work}

This paper presented a multiple lane detection algorithm. The process of the disparity estimation is optimised by using the disparity information acquired at time $t_{n}$ to suggest the search range at time $t_{n+1}$ ($n\geq0$). This not only improved the accuracy of the estimated disparities but also reduced the runtime of the algorithm. The runtime of the disparity estimation is reduced by approximately $37\%$. The lanes were detected using our previously published algorithm in \cite{Ozgunalp2017, Fan2018a}.  The experimental results illustrated that the proposed system works robustly and precisely for both highway and urban scenes and a $99\%$ successful detection rate was achieved when processing the KITTI dataset. 

However, some actual road conditions may result in failed detections. Therefore, we plan to train a deep neural network for dense vanishing point estimation. Furthermore, we plan to implement the proposed algorithm on some state-of-the-art embedded systems, such as Jetson TX2 for real-time purposes. 

\section{Acknowledgements}
This work is supported by the Research Grant Council of Hong Kong SAR Government, China, under Project No. 11210017, No. 16212815 and No. 21202816, NSFC U1713211 and Shenzhen Science, Technology and Innovation Commission (SZSTI) JCYJ20160428154842603, awarded to Prof. Ming Liu; Shenzhen Science, Technology and Innovation Commission (SZSTI) JCYJ20170818153518789 and National Natural Science Foundation of China No. 61603376, awarded to Dr. Lujia Wang.

\bibliographystyle{IEEEbib}

\begin{thebibliography}{10}
	
	\bibitem{Brink2017}
	James~A Brink, Ronald~L Arenson, Thomas~M Grist, Jonathan~S Lewin, and Dieter
	Enzmann,
	\newblock ``Bits and bytes: the future of radiology lies in informatics and
	information technology,''
	\newblock {\em European Radiology}, pp. 1--5, 2017.
	
	\bibitem{Fan2017}
	Rui Fan and Naim Dahnoun,
	\newblock ``Real-time implementation of stereo vision based on optimised
	normalised cross-correlation and propagated search range on a gpu,''
	\newblock in {\em Imaging Systems and Techniques (IST), 2017 IEEE International
		Conference on}. IEEE, 2017, pp. 241--246.
	
	\bibitem{Fan2016}
	Rui Fan, Victor Prokhorov, and Naim Dahnoun,
	\newblock ``Faster-than-real-time linear lane detection implementation using
	soc dsp tms320c6678,''
	\newblock in {\em Imaging Systems and Techniques (IST), 2016 IEEE International
		Conference on}. IEEE, 2016, pp. 306--311.
	
	\bibitem{Fan2018}
	R.~Fan, X.~Ai, and N.~Dahnoun,
	\newblock ``Road surface {3D} reconstruction based on dense subpixel disparity
	map estimation,''
	\newblock {\em IEEE Transactions on Image Processing}, vol. 27, no. 6, pp.
	3025--3035, June 2018.
	
	\bibitem{Fan2018c}
	Rui Fan, Yanan Liu, Xingrui Yang, Mohammud~Junaid Bocus, Naim Dahnoun, and
	Scott Tancock,
	\newblock ``Real-time stereo vision for road surface 3-d reconstruction,''
	\newblock {\em arXiv preprint arXiv:1807.07433}.
	
	\bibitem{Fan2018a}
	Rui Fan and Naim Dahnoun,
	\newblock ``Real-time stereo vision-based lane detection system,''
	\newblock {\em Measurement Science and Technology}, vol. 29, no. 7, 2018.
	
	\bibitem{Evans2018a}
	Meghan Evans, Rui Fan, and Naim Dahnoun,
	\newblock ``Iterative roll angle estimation from dense disparity map,''
	\newblock in {\em 2018 7th Mediterranean Conference on Embedded Computing
		(MECO)}. IEEE, 2018.
	
	\bibitem{Fan2018d}
	Rui Fan, Mohammud~Junaid Bocus, and Naim Dahnoun,
	\newblock ``A novel disparity transformation algorithm for road segmentation,''
	\newblock {\em Information Processing Letters}, vol. 140, pp. 18--24, 2018.
	
	\bibitem{Bertozzi1998}
	Massimo Bertozzi and Alberto Broggi,
	\newblock ``Gold: A parallel real-time stereo vision system for generic
	obstacle and lane detection,''
	\newblock {\em IEEE transactions on image processing}, vol. 7, no. 1, pp.
	62--81, 1998.
	
	\bibitem{Wang2004}
	Yue Wang, Eam~Khwang Teoh, and Dinggang Shen,
	\newblock ``Lane detection and tracking using b-snake,''
	\newblock {\em Image and Vision computing}, vol. 22, no. 4, pp. 269--280, 2004.
	
	\bibitem{Kluge1995}
	Karl Kluge and Sridhar Lakshmanan,
	\newblock ``A deformable-template approach to lane detection,''
	\newblock in {\em Intelligent Vehicles' 95 Symposium., Proceedings of the}.
	IEEE, 1995, pp. 54--59.
	
	\bibitem{Wang2008}
	Yan Wang, Li~Bai, and Michael Fairhurst,
	\newblock ``Robust road modeling and tracking using condensation,''
	\newblock {\em IEEE Transactions on Intelligent Transportation Systems}, vol.
	9, no. 4, pp. 570--579, 2008.
	
	\bibitem{Zhou2006}
	Yong Zhou, Rong Xu, Xiaofeng Hu, and Qingtai Ye,
	\newblock ``A robust lane detection and tracking method based on computer
	vision,''
	\newblock {\em Measurement science and technology}, vol. 17, no. 4, pp. 736,
	2006.
	
	\bibitem{Kreucher1999}
	Chris Kreucher and Sridhar Lakshmanan,
	\newblock ``Lana: a lane extraction algorithm that uses frequency domain
	features,''
	\newblock {\em IEEE Transactions on Robotics and automation}, vol. 15, no. 2,
	pp. 343--350, 1999.
	
	\bibitem{Jung2005}
	Cl{\'a}udio~Rosito Jung and Christian~Roberto Kelber,
	\newblock ``An improved linear-parabolic model for lane following and curve
	detection,''
	\newblock in {\em Computer Graphics and Image Processing, 2005. SIBGRAPI 2005.
		18th Brazilian Symposium on}. IEEE, 2005, pp. 131--138.
	
	\bibitem{Wang2000}
	Yue Wang, Dinggang Shen, and Eam~Khwang Teoh,
	\newblock ``Lane detection using spline model,''
	\newblock {\em Pattern Recognition Letters}, vol. 21, no. 8, pp. 677--689,
	2000.
	
	\bibitem{Nieto2007}
	Marcos Nieto, Luis Salgado, Fernando Jaureguizar, and Julian Cabrera,
	\newblock ``Stabilization of inverse perspective mapping images based on robust
	vanishing point estimation,''
	\newblock in {\em Intelligent Vehicles Symposium, 2007 IEEE}. IEEE, 2007, pp.
	315--320.
	
	\bibitem{Schreiber2005}
	David Schreiber, Bram Alefs, and Markus Clabian,
	\newblock ``Single camera lane detection and tracking,''
	\newblock in {\em Intelligent Transportation Systems, 2005. Proceedings. 2005
		IEEE}. IEEE, 2005, pp. 302--307.
	
	\bibitem{Hanwell2012}
	David Hanwell and Majid Mirmehdi,
	\newblock ``Detection of lane departure on high-speed roads.,''
	\newblock in {\em ICPRAM (2)}, 2012, pp. 529--536.
	
	\bibitem{Fardi2004}
	Basel Fardi and Gerd Wanielik,
	\newblock ``Hough transformation based approach for road border detection in
	infrared images,''
	\newblock in {\em Intelligent Vehicles Symposium, 2004 IEEE}. IEEE, 2004, pp.
	549--554.
	
	\bibitem{Ozgunalp2017}
	Umar Ozgunalp, Rui Fan, Xiao Ai, and Naim Dahnoun,
	\newblock ``Multiple lane detection algorithm based on novel dense vanishing
	point estimation,''
	\newblock {\em IEEE Transactions on Intelligent Transportation Systems}, vol.
	18, no. 3, pp. 621--632, 2017.
	
	\bibitem{Labayrade2002}
	Raphael Labayrade, Didier Aubert, and J-P Tarel,
	\newblock ``Real time obstacle detection in stereovision on non flat road
	geometry through" v-disparity" representation,''
	\newblock in {\em Intelligent Vehicle Symposium, 2002. IEEE}. IEEE, 2002,
	vol.~2, pp. 646--651.
	
	\bibitem{Leu2011}
	Stefan Leutenegger, Margarita Chli, and Roland~Y Siegwart,
	\newblock ``Brisk: Binary robust invariant scalable keypoints,''
	\newblock in {\em 2011 International conference on computer vision}. IEEE,
	2011, pp. 2548--2555.
	
	\bibitem{Fan2018b}
	Rui Fan, Yanan Liu, Mohammud~Junaid Bocus, Lujia Wang, and Ming Liu,
	\newblock ``Real-time subpixel fast bilateral stereo,''
	\newblock {\em arXiv preprint arXiv:1807.02044}.
	
	\bibitem{Hillel2014}
	Aharon~Bar Hillel, Ronen Lerner, Dan Levi, and Guy Raz,
	\newblock ``Recent progress in road and lane detection: a survey,''
	\newblock {\em Machine vision and applications}, vol. 25, no. 3, pp. 727--745,
	2014.
	
\end{thebibliography}

\end{document}